\documentclass[10pt,twocolumn,letterpaper]{article}

\usepackage[pagenumbers]{cvpr} 

\usepackage[dvipsnames]{xcolor}

\usepackage{graphicx}
\usepackage{amsmath}
\usepackage{amssymb}
\usepackage{booktabs}
\usepackage{caption}
\usepackage{multirow}

\definecolor{cvprblue}{rgb}{0.21,0.49,0.74}
\usepackage[pagebackref,breaklinks,colorlinks,citecolor=cvprblue]{hyperref}

\usepackage[capitalize]{cleveref}
\crefname{section}{Sec.}{Secs.}
\Crefname{section}{Section}{Sections}
\Crefname{table}{Table}{Tables}
\crefname{table}{Tab.}{Tabs.}
\usepackage{bm}
\usepackage{capt-of}
\usepackage{cuted}
\usepackage{multirow}
\usepackage{relsize}

\usepackage[normalem]{ulem}

\def\paperID{5996} 
\def\confName{CVPR}
\def\confYear{2024}

\title{Smooth Diffusion: Crafting Smooth Latent Spaces in Diffusion Models}

\author{
  Jiayi Guo$^{1,2}$\thanks{Equal contribution.},
  Xingqian Xu$^{1,3}$\footnotemark[1],
  Yifan Pu$^{2}$,
  Zanlin Ni$^{2}$,
  Chaofei Wang$^{2}$,
  Manushree Vasu$^{1}$,\\ 
    Shiji Song$^{2}$,
    Gao Huang$^{2\dagger}$, 
  Humphrey Shi$^{1,3}$\thanks{Corresponding authors.}\\
  {\small
    $^{1}$SHI Labs @ Georgia Tech \& UIUC\ \ \
    $^{2}$Tsinghua University\ \ \
    $^{3}$Picsart AI Research (PAIR)}\\
     {\small \textbf{\url{https://github.com/SHI-Labs/Smooth-Diffusion}}}
  \vspace{-16mm}
}

\begin{document}

\maketitle

\begin{strip}
    \centering
  \includegraphics[width=0.99\linewidth]{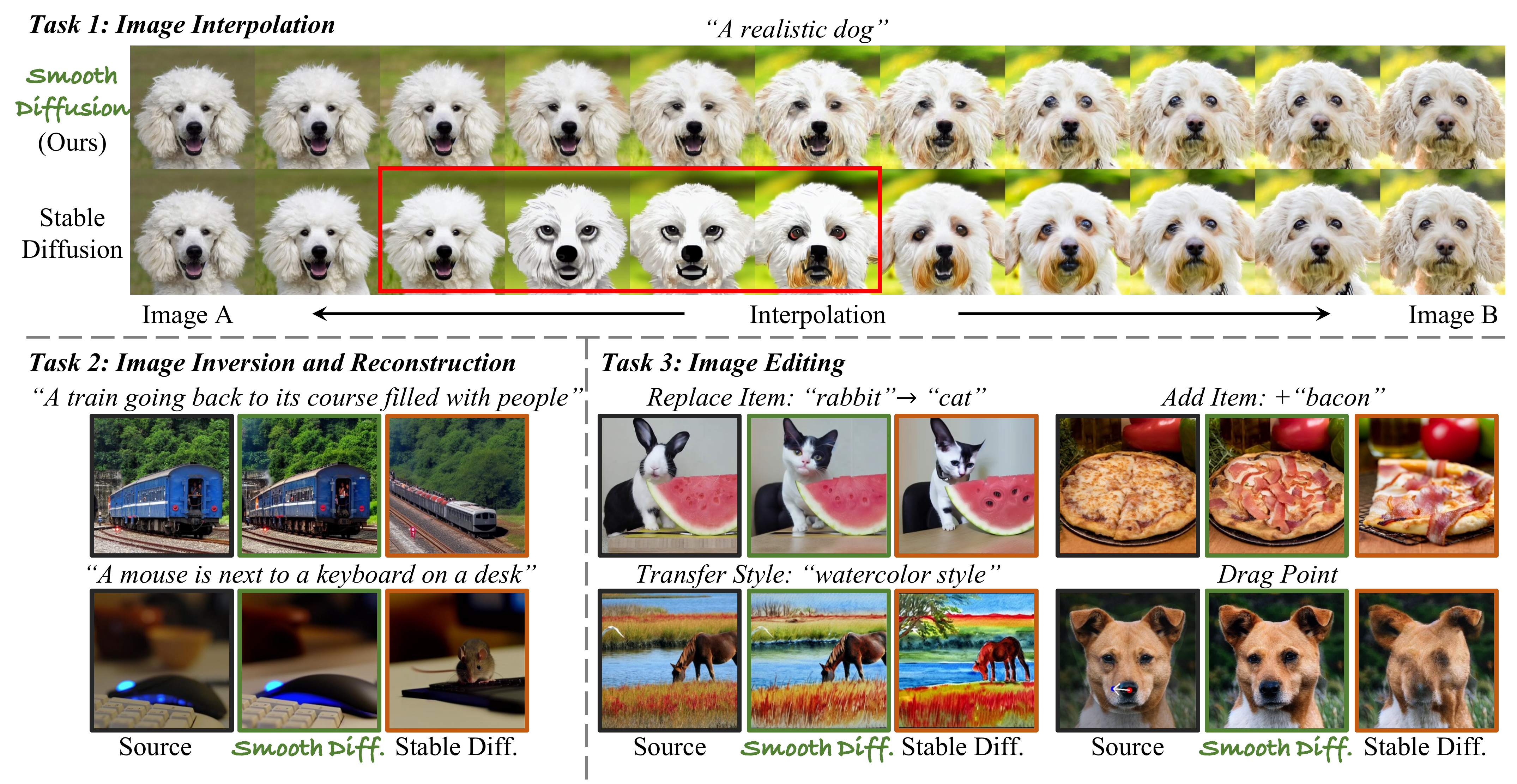}
    \vspace{-2mm}   
  \captionof{figure}{\textbf{Smooth Diffusion for downstream image synthesis tasks.} Our method formally introduces \textcolor{blue}{\textbf{latent space smoothness}} to diffusion models like Stable Diffusion~\cite{sd}. This smoothness dramatically aids various tasks in: 1) improving continuity of transitions in image interpolation, 2) reducing approximation errors in image inversion, \& 3) better preserving unedited contents in image editing.}
  \label{fig:fig1}
\end{strip}


\begin{abstract}


Recently, diffusion models have made remarkable progress in text-to-image (T2I) generation, synthesizing images with high fidelity and diverse contents. Despite this advancement, \textbf{latent space smoothness} within diffusion models remains largely unexplored. Smooth latent spaces ensure that a perturbation on an input latent corresponds to a steady change in the output image. This property proves beneficial in downstream tasks, including image interpolation, inversion, and editing. In this work, we expose the non-smoothness of diffusion latent spaces by observing noticeable visual fluctuations resulting from minor latent variations. To tackle this issue, we propose \textbf{Smooth Diffusion}, a new category of diffusion models that can be simultaneously high-performing and smooth. Specifically, we introduce \textbf{Step-wise Variation Regularization} to enforce the proportion between the variations of an arbitrary input latent and that of the output image is a constant at any diffusion training step. In addition, we devise an interpolation standard deviation (ISTD) metric to effectively assess the latent space smoothness of a diffusion model. Extensive quantitative and qualitative experiments demonstrate that Smooth Diffusion stands out as a more desirable solution not only in T2I generation but also across various downstream tasks. 
Smooth Diffusion is implemented as a plug-and-play Smooth-LoRA to work with various community models. 
Code is available at 
\href{https://github.com/SHI-Labs/Smooth-Diffusion}{https://github.com/SHI-Labs/Smooth-Diffusion}.

\vspace{-2mm}
\end{abstract}

\section{Introduction}
\label{sec:intro}

In recent years, diffusion models~\cite{ddpm, diffusion-beats-gans, sd} have rapidly grown into very powerful tools for generative AI, particularly for text-to-image generation. The remarkable ability of diffusion models, generating high-quality photorealistic images from open-book contexts, has been highlighted in many research and commercial products. Such success has also inspired various diffusion-based downstream tasks, including image interpolation~\cite{sdw,anid}, inversion~\cite{ddim,nti,edict,pti,aidi},  editing~\cite{sdedit,p2p,pnp, disentanglement,pix2pixzero,cyclediffusion,dragdiffusion,dragondiffusion}, \etc.

Despite the great success in the generation field, diffusion models occasionally produce low-quality results with undesirable and unpredictable behaviors.
Specifically speaking, for image interpolation, the Stable Diffusion Walk (SDW)~\cite{sdw} test examines latent space with spherical linear interpolations, usually resulting in highly fluctuated outputs with unpredictable visual appearance. Examples can be found in~\cref{fig:fig1} Task 1, in which such interpolation exhibits undesired sharp changes as well as ``cartoon-ization'' on photorealistic dog images, highlighted in the red box. For the image inversion task shown in~\cref{fig:fig1} Task 2, a naive application of DDIM inversion~\cite{ddim} cannot reconstruct images faithfully from the sources. Instead, it generates incorrect colors and object orientations, and misinterprets the computer mouse as an animal mouse. For the image editing task shown in~\cref{fig:fig1} Task 3, one may notice that only minor text prompt editing can lead to major updates on image contents and layouts, in which the object (\ie the cat's pose, the horse's location, the shape of the pizza) can be wildly and incorrectly altered. Moreover, current diffusion models are unsuited to drag-based editing~\cite{dragdiffusion} because a fine-engineered drag method still has a noticeably large chance of breaking objects' shape and semantics.

In this work, we step into an important but under-explored area: to improve the latent space smoothness of diffusion models. Our motivation to enhance latent smoothness comes from the real-world demand to improve the output qualities of the aforementioned downstream tasks.
A smooth latent space implies a robust visual variation under a minor latent change. Therefore, enhancing such smoothness could help improve the continuity of image interpolation, expand the capacity of image inversion, and maintain correct semantics in image editing. Notably, prior works in GANs~\cite{sg2,sg3,interfacegan} have demonstrated that the smooth latent space of the generator can significantly improve downstream tasks' quality, offering additional evidence of the importance of this area.

To achieve our goal, we propose \textbf{Smooth Diffusion}, a new category of diffusion models that can be simultaneously high-performing and smooth. We start our exploration by first formalizing the objective for Smooth Diffusion, in which fixed-size perturbations $\Delta\bm\epsilon$ on a latent noise $\bm\epsilon$ should produce smooth visual changes $\Delta\widehat{\bm{x}_0}$ on the synthetic image $\widehat{\bm{x}_0}$, rounded to a constant ratio $C$. Although one may think that according to the formulation, the smoothness constraint could be an accessible train-time loss. Actually, there is no direct application of such regularization from inference to training, and the challenge lies in the fact that in each training iteration (\ie, back-propagation), diffusion models optimize only a ``$t$-step snapshot'' instead of the entire $T$-step diffusion process.

Therefore, we introduce \textbf{Step-wise Variation Regularization}, a novel regularization that seamlessly incorporates our Smooth Diffusion's inference-time objective to training. This regularization aims to bound the 2-norm of output variation $\Delta\widehat{\bm{x}_0}$ given a fixed-size change $\Delta\bm{x}_t$ in input $\bm{x}_t$ at an arbitrary step $t$. The rationale of the reformulation is intuitive: If $\bm{x}_t$ and $\widehat{\bm{x}_0}$ exhibit smooth changes at any $t$, then the relation between the latent noise $\bm\epsilon$ (\ie $\bm{x}_T$) and $\widehat{\bm{x}_0}$ is just the accumulation of smooth variations and thus can be smooth as well. More details can be found in~\cref{sec:3}.

In practice, our Smooth Diffusion is trained on top of a well-known text-to-image model: Stable Diffusion~\cite{sd}. We examine and demonstrate that 
Smooth Diffusion dramatically improves the latent space smoothness over its baseline. Meanwhile, we conduct extensive research across numerous downstream tasks, including but not limited to image interpolation, inversion, editing, \etc. Both qualitative and quantitative results support our conclusion that Smooth Diffusion can be the next-gen high-performing generative model not only for the baseline text-to-image task but across various downstream tasks. 

\section{Related Work}
\label{sec:related}

\textbf{Diffusion models} are initiated from a family of prior works including but not limited to~\cite{dm_early1, mcm0, mcm1, dm_early0}. Since then, DDPM~\cite{ddpm} introduced an image-based noise prediction model, becoming one of the most popular image generation research. Later works~\cite{improved-diffusion, ddim, diffusion-beats-gans} extended DDPM, demonstrating that diffusion models perform on-par and even surpass GAN-based methods~\cite{gan, sg, sg2, sgada, sg3}.
Recently, generating images from text prompts (T2I) become an emerging field, among which diffusion models~\cite{glide, dalle2, imagen, vqdiffusion, sd} have become quite visible to the public. For example, Stable Diffusion (SD)~\cite{sd} consists of VAE~\cite{vae} and CLIP~\cite{clip}, diffuses latent space, and yields an outstanding balance between quality and speed. Following SD~\cite{sd}, researchers also explored diffusion approaches for controls such as ControlNet~\cite{controlnet, t2iadapter, zhang2023forget, composer, pairdiffusion, promptdiffusion, refpaint, unicontrol, unicontrol2, pfd, ip} and multimodal such as Versatile Diffusion~\cite{vd,codi, unibrain, unidiffuser}. Works from a different track reduce diffusion steps to improve speed~\cite{zhang2022fast,analyticdpm, dpmslover, edm, unipc, progressive, guide-distill, consistency}, or restrict data and domain for few-shot learning~\cite{dreambooth, lora, specdiff, ipl}, all had successfully maintained a high output quality.

\vspace{0.2cm}
\noindent\textbf{Smooth latent space} was one of the prominent properties of SOTA GAN works~\cite{biggan, sg2, sgada, sg3}, while exploring such property went through the decade-long GAN research~\cite{gan, gan-base2}, whose goals were mainly robust training. Ideas such as Wasserstein GAN~\cite{wgan, wgan-gp} had proved to be effective, which enforced the Lipschitz continuity on discriminator via gradient penalties. Another technique, namely path length regularization, related to the Jacobian clamping in~\cite{is-gan-perform}, was adapted in StyleGAN2~\cite{sg2} and later became a standard setting for GAN-based generators~\cite{comodgan, shgan, infinitygan, pigan}. Benefiting from the smoothness property, researchers managed to manipulate latent space in many downstream research projects. Works such as~\cite{gan-dissection, xia2020controllable, faceid, interfacegan} explored latent space disentanglement. GAN-inverse~\cite{pan2021exploiting, image2stylegan++, restyle, gan-inversion} had also proved to be feasible, along with a family of image editing approaches~\cite{in-domain-gan, pivotal-tuning-gan, psp, e4e, styleclip, draggan}. As aforementioned, our work aims to investigate the latent space smoothness for diffusion models, which by far remains unexplored.


\vspace{0.2cm}
\section{Methodology}\label{sec:3}
\label{sec:method}

In this section, we first introduce preliminaries of our method, including diffusion process~\cite{ddpm}, diffusion inversion~\cite{ddim, diffusion-beats-gans, nti} and low-rank adaptation~\cite{lora} (\cref{sec:pre}). Then Smooth Diffusion is proposed with its definition, objective (\cref{sec:sd}) and regularization function (\cref{sec:3-3}).


\vspace{0.1cm}
\subsection{Preliminaries}\label{sec:pre}
\vspace{0.1cm}

\textbf{Diffusion process}~\cite{ddpm} is a kind of Markov chain that gradually adds random noise $\bm{\epsilon}_t \sim N(\bm{0}, \bm{I})$ to ground truth signal $\bm{x}_0 \sim p(\bm{x}_0)$, making $\bm{x}_T$ in a total of ${T}$ steps. At each step, The noisy data $\bm{x}_t$ is computed as:

\begin{equation}
\setlength{\abovedisplayskip}{-8pt}
\setlength{\belowdisplayskip}{6pt}
\begin{aligned}
    \bm{x}_t &= \sqrt{1-\beta_t}\bm{x}_{t-1} + \sqrt{\beta_t} \bm{\epsilon}_t,\quad t =1,2,\cdots,T,
\end{aligned}
\label{eq:xt}
\end{equation}

\noindent where $\beta_t$ is the preset diffusion rate at step $t$. By making $\alpha_t = 1-\beta_t$, $\overline{\alpha_t} = \prod_{t=1}^{T}\alpha_t$ and $\bm{\epsilon} \sim N(\bm{0}, \bm{I})$, we have the following equivalents:

\begin{equation}
\setlength{\abovedisplayskip}{-3pt}
\setlength{\belowdisplayskip}{6pt}
\begin{aligned}
    \bm{x}_t &= \sqrt{\alpha_t}\bm{x}_{t-1} + \sqrt{1-\alpha_t} \bm{\epsilon}_t\\
    & = \sqrt{\overline{\alpha_t}}\bm{x}_{0} + \sqrt{1-\overline{\alpha_t}} \bm{\epsilon},\quad t =1,2,\cdots,T.
\end{aligned}
\label{eq:xt2}
\end{equation}

 A diffusion model $\epsilon_\theta(\bm{x}_t, t)$ is then trained to estimate $\bm{\epsilon}_t$ from $\bm{x}_t$, by which one can predict the original signal $\bm{x}_0$ by gradually remove noise from the degraded $\bm{x}_T$~\cite{ddim}. This is commonly known as the backward diffusion process:

\begin{equation}
\setlength{\abovedisplayskip}{-6pt}
\setlength{\belowdisplayskip}{0pt}
{
\begin{aligned}
\widehat{\bm{x}_{t-1}} = \sqrt{\frac{\alpha_{t-1}}{\alpha_{t}}}\widehat{\bm{x}_{t}}+\left(\sqrt{\frac{1}{\alpha_{t-1}}-1}-\sqrt{\frac{1}{\alpha_t}-1}\right)\cdot{\epsilon_\theta}(\widehat{\bm{x}_t}, t).
\label{eq:ddim}
\end{aligned}
}
\end{equation}

\noindent\textbf{Diffusion inversion}~\cite{ddim, diffusion-beats-gans, nti} targets to recover the exact backward diffusion process (\ie $\widehat{\bm{x}_t}, \epsilon_\theta(\widehat{\bm{x}_t},t), t=1,...,T$) from a known final prediction $\widehat{\bm{x}_0}$. One of the common technique for such inversion is \textit{DDIM inversion}~\cite{ddim, diffusion-beats-gans}, which reverses~\cref{eq:ddim} under a local linear approximation:

\begin{equation}
\setlength{\abovedisplayskip}{-3pt}
\setlength{\belowdisplayskip}{0pt}
{
\begin{aligned}
\widetilde{\bm{x}_{t+1}} = 
    \sqrt{\frac{\alpha_{t+1}}{\alpha_{t}}}\widetilde{\bm{x}_{t}} + 
    \left(
        \sqrt{\frac{1}{\alpha_{t+1}}-1} - 
        \sqrt{\frac{1}{\alpha_t}-1}
    \right) \cdot {\epsilon_\theta}(\widetilde{\bm{x}_t}, t),
\label{eq:ddimin}
\end{aligned}
}
\end{equation}

\noindent where $\widetilde{\bm{x}_t}$ represent the estimated $\widehat{\bm{x}_t}$ at time $t$. However, DDIM inversion is only a rough estimation. For text-to-image diffusion, a more advanced technique, \textit{Null-Text Inversion}~\cite{nti}, optimizes additional null-text embeddings $\{\varnothing_t\}_{t=1}^{T}$ for each step $t$, simulating the backward process with ${\epsilon_\theta}(\bm{x}_t, t, \xi, \varnothing_t)$, where $\xi$ is the input text embedding. The predicted null-text $\varnothing_t$ is the null input of the classifier-free guidance~\cite{cfg} with a guidance scale $w$:

\begin{equation}
\setlength{\abovedisplayskip}{0pt}
\setlength{\belowdisplayskip}{0pt}
{
\begin{aligned}
{\epsilon_\theta}(\bm{x}_t, t, \xi, \varnothing_t) = 
w\cdot{\epsilon_\theta}(\bm{x}_t, t, \xi) 
+ (1-w)\cdot{\epsilon_\theta}(\bm{x}_t, t, \varnothing_t).
\label{eq:cfg}
\end{aligned}
}\end{equation}

\noindent\textbf{Low-rank adaptation (LoRA)}~\cite{lora} is initially proposed to efficiently adapt large pretrained models to downstream tasks. The key assumption of LoRA is that the weight changes required during adaptation maintain a low rank. Given a pretrained model weight $W_0\in\mathbb{R}^{d\times k}$, its updated weight $\Delta W$ is expressed as a low rank decomposition:
\begin{equation}
\setlength{\abovedisplayskip}{5pt}
\setlength{\belowdisplayskip}{5pt}
{
\begin{aligned}
    W_0 + \Delta W =  W_0 + BA,
\end{aligned}
}\end{equation}
where $B\in\mathbb{R}^{d\times r}$, $A\in\mathbb{R}^{r\times k}$ and $r\ll \min(d,k)$. During adaptation, $W_0$ is frozen, while $B$ and $A$ are trainable.

\begin{figure*}[t]
  \centering
  \includegraphics[width=0.95\linewidth]{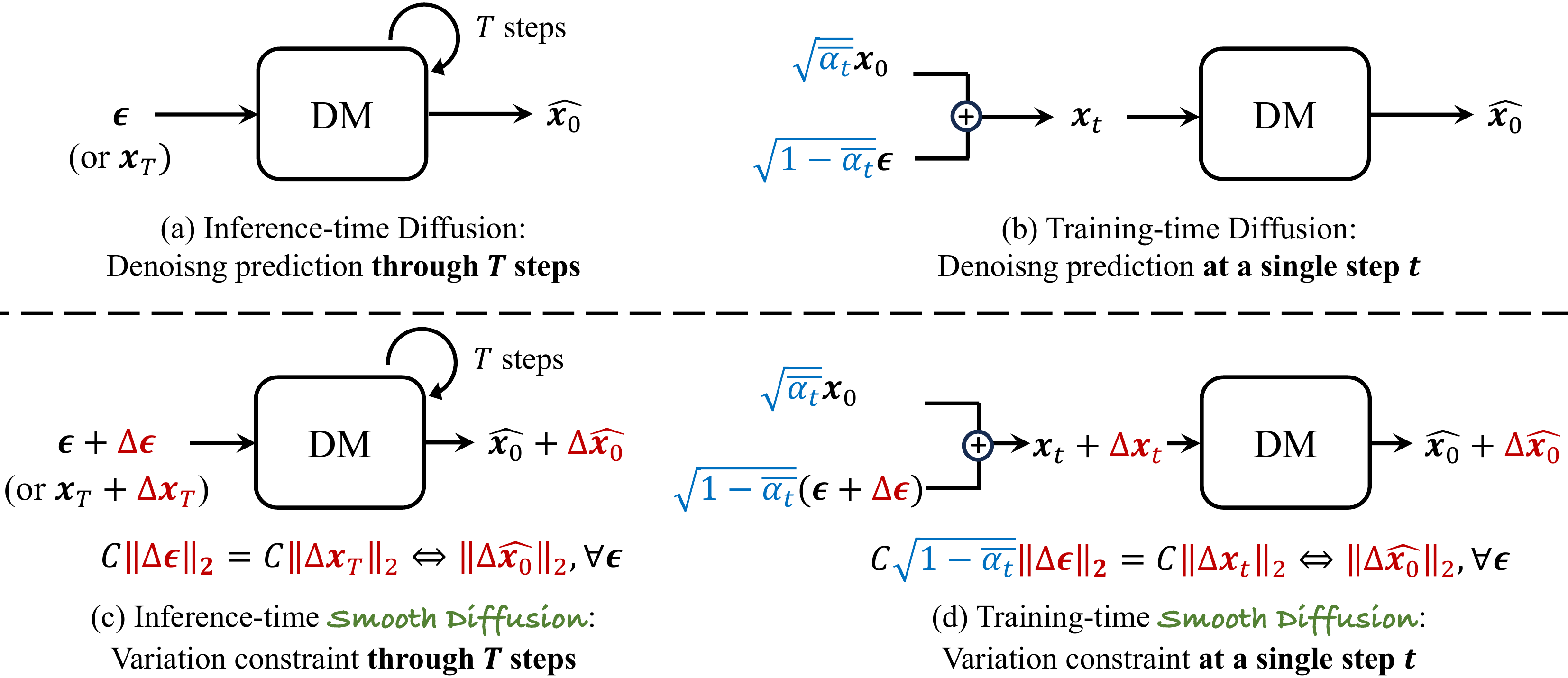}
  \caption{\textbf{Illustration of Smooth Diffusion.} Smooth Diffusion (c) enforces the ratio between the variation of the input latent ($\Vert\Delta \bm{\epsilon}\Vert_2$ or $\Vert\Delta \bm{x}_T\Vert_2$) and the variation of the output prediction ($\Vert\Delta\widehat{\bm{x}_0}\Vert_2$) is a constant $C$. 
  Training-time Diffusion (b) optimizes a ``$t$-step snapshot'' of the denoising prediction process in Inference-time Diffusion (a). Similarly, we propose Training-time Smooth Diffusion (d) to optimize a ``$t$-step snapshot'' of the variation constraint in Inference-time Smooth Diffusion (c). DM: Diffusion model.}
  \label{fig:method}
  \vspace{-2mm}
\end{figure*}

\vspace{0.1cm}
\subsection{Smooth Diffusion}\label{sec:sd}
\vspace{0.1cm}

As previously mentioned, modern diffusion models (DM) do not guarantee latent space smoothness, creating not only research gaps between GANs and diffusions but also unexpected challenges in downstream tasks. To address these issues, we propose \textbf{Smooth Diffusion}, a novel class of high-performing diffusion models with enhanced smoothness over its latent space. The underlining of Smooth Diffusion is the newly proposed training scheme in which we carried out a \textbf{Step-wise Variation Regularization} to enhance model smoothness.

To better explain our aims, we adopt the same terminologies from the standard inference-time diffusion process (\cref{fig:method}a), involving a $T$ steps procedure that transforms the random noise $\bm{\epsilon}$ (\ie, $\bm{x}_T$) to the prediction $\widehat{\bm{x}_0}$. The overall objective of Smooth Diffusion can then be written in Eq.~\ref{eq:inference_goal}: in which we expect that a fixed-size change $\Delta \bm{\epsilon}$ on $\bm{\epsilon}$ (\ie, $\Delta \bm{x}_T$ on $\bm{x}_T$) will finally lead to a non-zero, fixed-size change $\Delta \widehat{\bm{x}_0}$ on $\widehat{\bm{x}_0}$, up to a constant ratio $C$: 

\vspace{-0.2cm}
\begin{equation}
\setlength{\abovedisplayskip}{5pt}
\setlength{\belowdisplayskip}{5pt}{
\begin{aligned}
    \Vert\Delta\widehat{\bm{x}_0}\Vert_2 
    \Leftrightarrow C\Vert\Delta \bm{x}_T\Vert_2 
    = C\Vert\Delta \bm{\epsilon}\Vert_2
    ,\ \forall \bm{\epsilon},
\end{aligned}}
\label{eq:inference_goal}
\end{equation}

\noindent Notice that by definition, $\bm{x}_T$ is the initial input of the backward diffusion loop in Eq.~\ref{eq:ddim}. Since $\bm{x}_T$ is close to $\bm{\epsilon} \sim N(\bm{0}, \bm{1})$, for simplicity, we make them equivalent in all the following equations. 

Nevertheless, one may notice that our inference-time objective in Eq.~\ref{eq:inference_goal} cannot be directly transformed into a training loss function. This is because, in one training iteration (\ie, back-propagation), diffusion models optimize only a ``$t$-step snapshot'' of the diffusion process (\cref{fig:method}b), where $t$ is uniformly sampled from 1 to $T$. Hence, the proposed ``global'' objective (Eq.~\ref{eq:inference_goal}) for the entire $T$-step process is not accessible in training. Therefore, we need to reformulate our global objective into a \textbf{step-wise objective} shown in Eq.~\ref{eq:train_goal}, which can later be integrated into the diffusion training process as a loss function: 
\begin{equation}
\setlength{\abovedisplayskip}{6pt}
\setlength{\belowdisplayskip}{6pt}{
\begin{aligned}
\Vert \Delta\widehat{\bm{x}_0} \Vert_2 
\Leftrightarrow C \Vert \Delta\bm{x}_t \Vert_2
= C 
    \sqrt{1-\overline{\alpha_t}}
    \Vert\Delta\bm\epsilon 
\Vert_2, 
\ \forall \bm{\epsilon},
\end{aligned}}\label{eq:train_goal}\end{equation}
where $C$ is a non-zero constant. This step-wise objective indicates that at each training step, variations $\Delta \bm{\epsilon}$ on $\bm{\epsilon}$ should imply variations $\Delta\bm{x}_t$ on $\bm{x}_t$ with a ratio proportional to $\sqrt{1-\overline{\alpha_t}}$. The rationale of Eq.~\ref{eq:train_goal} is intuitive: If $\bm{x}_t$ and $\widehat{\bm{x}_0}$ show smooth changes at any $t$, then the relation between the latent noise $\bm\epsilon$ (\ie $\bm{x}_T$) and $\widehat{\bm{x}_0}$ is just the accumulation of smooth variations and thus can be smooth as well. 

\subsection{Step-wise Variation Regularization}\label{sec:3-3}


While the motivation and formulation of the Smooth Diffusion objective are presented, how to realize such an objective remains unexplained. Therefore, in this section, we introduce \textbf{Step-wise Variation Regularization} to effectively integrate the step-wise objective into diffusion training.



We draw inspiration from the regularization techniques~\cite{is-gan-perform, sg2} adopted in GAN training. The core idea of Step-wise Variation Regularization is to bound the Jacobian matrix $\mathbf{J}_{\bm\epsilon} = \partial\widehat{\bm{x}_0} / \partial\bm\epsilon$ of the diffusion system by minimizing the following regularization loss at any $\bm{x}_0, \bm\epsilon,$ and step $t$:
\begin{equation}
\setlength{\abovedisplayskip}{5pt}
\setlength{\belowdisplayskip}{5pt}{
\begin{aligned}
    \mathcal{L}_{\rm{reg}} 
    = \mathbb{E}_{
        \Delta\widehat{\bm{x}_0},
        \bm\epsilon
    }
    \left(
        \sqrt{1-\overline{\alpha_t}}
        \Vert 
            \mathbf{J}_{\bm\epsilon}^{\rm T}\Delta \widehat{\bm{x}_0}
        \Vert_2 -a
    \right)^2,
\end{aligned}}\label{eq:reg_diffusion}\end{equation}

\noindent where $\Delta\widehat{\bm{x}_0}$ is the normally sampled pixel intensities normalized to unit length, $\bm\epsilon$ is a normally sampled noise in Eq.~\ref{eq:xt2}, and $a$ is the exponential moving average of $\sqrt{1-\overline{\alpha_t}}\Vert \mathbf{J}_{\bm\epsilon}^{\rm T}\Delta \widehat{\bm{x}_0}\Vert_2$ computed online during training. In practice, we compute Eq.~\ref{eq:reg_diffusion} via standard backpropagation with the following identity:
\begin{equation}
\setlength{\abovedisplayskip}{5pt}
\setlength{\belowdisplayskip}{5pt}{\begin{aligned}
    \sqrt{1-\overline{\alpha_t}}
    \Vert 
        \mathbf{J}_{\bm\epsilon}^{\rm T}\Delta \widehat{\bm{x}_0}
    \Vert_2
    = 
    \Vert
        \nabla_{\bm\epsilon}(
            \sqrt{1-\overline{\alpha_t}}
            \widehat{\bm{x}_0} \cdot \Delta\widehat{\bm{x}_0})
    \Vert_2.
\end{aligned}}\label{eq:bypass_jacobian}\end{equation}
The identity holds since $\Delta \widehat{\bm{x}_0}$ is independently sampled, and uncorrelated with $\bm\epsilon$.

Next, we prove that the proposed objective in Eq. \ref{eq:reg_diffusion} exactly matches our optimization goal in Eq. \ref{eq:train_goal}. One preliminary result, proven in~\cite{sg2}, is that in high dimensions, Eq. \ref{eq:reg_diffusion} is minimized when $\mathbf{J}_{\bm\epsilon}$ is orthogonal at any $\bm\epsilon$ up to a global scaling factor $\mathcal{K}$ (\ie $\mathbf{J}_{\bm\epsilon} \cdot \mathbf{J}_{\bm\epsilon}^{\rm T} = \mathcal{K} \cdot \bm{I}$). By applying the orthogonality of $\mathbf{J}_{\bm\epsilon}$, we have the following:
\begin{equation}\setlength{\abovedisplayskip}{5pt}
\setlength{\belowdisplayskip}{5pt}{\begin{aligned}
    \mathbf{J}_{\bm\epsilon}^{\rm T} \Delta \widehat{\bm{x}_0}
    = 
    \mathcal{K}\mathbf{J}_{\bm\epsilon}^{-1} \Delta \widehat{\bm{x}_0}
    =
    \mathcal{K} \frac
        {\partial \bm\epsilon}
        {\partial \widehat{\bm{x}_0}}
    \cdot \Delta \widehat{\bm{x}_0}
    =
    \mathcal{K} \Delta \bm\epsilon.
\end{aligned}}\label{eq:proof_stage1}\end{equation}
When $\mathcal{L}_{\rm{reg}}$ in Eq. \ref{eq:reg_diffusion} reaches its optimal, we then have:
\begin{equation}\setlength{\abovedisplayskip}{5pt}
\setlength{\belowdisplayskip}{5pt}{\begin{aligned}
    a 
    = 
    \sqrt{1-\overline{\alpha_t}}
    \Vert 
        \mathbf{J}_{\bm\epsilon}^{\rm T}\Delta \widehat{\bm{x}_0}
    \Vert_2
    = 
    \sqrt{1-\overline{\alpha_t}}
    \mathcal{K} \Vert \Delta \bm\epsilon \Vert_2.
\end{aligned}}\label{eq:proof_stage2}\end{equation}

\noindent Notice that $a=a\Vert\Delta\widehat{\bm{x}_0}\Vert_2$, since $\Vert\Delta\widehat{\bm{x}_0}\Vert_2=1$ is the aforementioned random unit length vector. Hence, we can finally reformulate the expression:
\begin{equation}\setlength{\abovedisplayskip}{5pt}
\setlength{\belowdisplayskip}{5pt}{\begin{aligned}
    \Vert
        \Delta\widehat{\bm{x}_0}
    \Vert_2
    &=
    \frac{\mathcal{K}}{a} 
    \sqrt{1-\overline{\alpha_t}}
    \Vert \Delta \bm\epsilon \Vert_2
    \\&=
    \mathcal{C} \sqrt{1-\overline{\alpha_t}}
    \Vert \Delta \bm\epsilon \Vert_2,
\end{aligned}}\label{eq:proof_stage3}\end{equation}
\noindent which exactly matches our proposed objective in Eq. \ref{eq:train_goal}. 

\begin{figure*}[t]
  \centering
  \includegraphics[width=\linewidth]{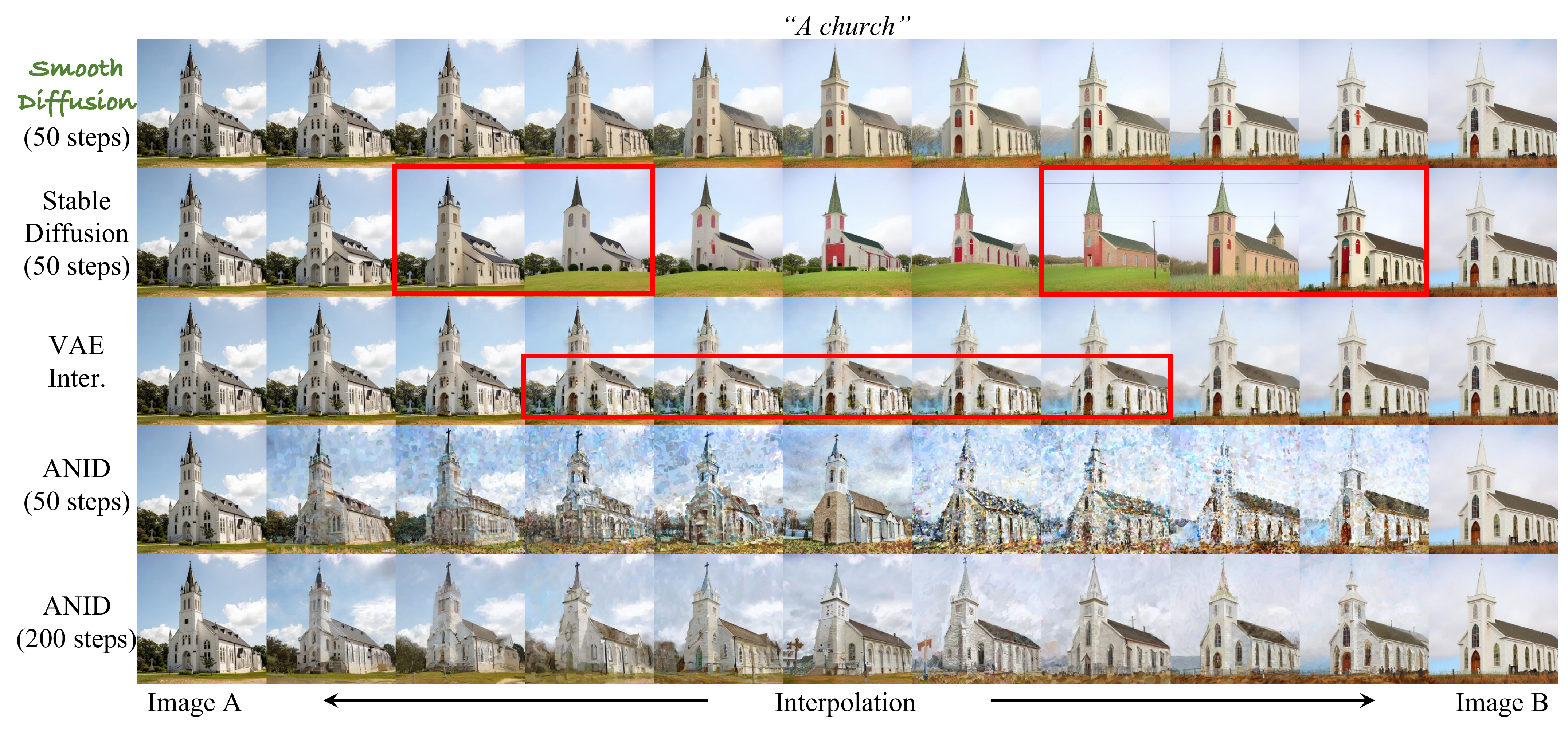}
    \vspace{-6mm}   
  \caption{\textbf{Image interpolation comparison results.} For Smooth Diffusion and Stable Diffusion~\cite{sd}, real images (Image A and B) are inverted into latents using NTI~\cite{nti}. We perform spherical linear interpolations between latents (also known as Stable Diffusion Walk~\cite{sdw}) and concatenate the resulting images as a transition sequence. VAE Inter. performs interpolations within the VAE space of Stable Diffusion. ANID~\cite{anid} first adds noise
to real images and subsequently denoises the interpolated noisy images using Stable Diffusion.
  }
  \label{fig:main}
  \vspace{-2mm}
\end{figure*}

To summarize, during training, the Smooth Diffusion objective encompasses a combination of $\mathcal{L}_{\rm{base}}$ and $\mathcal{L}_{\rm{reg}}$:
\begin{equation}
\setlength{\abovedisplayskip}{5pt}
\setlength{\belowdisplayskip}{5pt}{\begin{aligned}
\mathcal{L} = \mathcal{L}_{\rm{base}} + \lambda \mathcal{L}_{\rm{reg}}, \label{eq:finalloss}
\end{aligned}}\end{equation}
where $\mathcal{L}_{\rm{base}}$ denotes the basic training objective of a diffusion model and $\lambda$ represents a ratio parameter controlling the intensity of Step-wise Variation Regularization.

\section{Experiments}

\label{sec:exp}

\subsection{Experimental Setup}
\label{sec:setup}

\textbf{Baselines and settings.} We select the Stable Diffusion~\cite{sd} as the primary baseline for all tasks. Additionally, for image interpolation, we adopt a VAE-space interpolation and ANID~\cite{anid} as competitors. 
For image inversion, we integrate Smooth Diffusion and Stable Diffusion with DDIM inversion~\cite{ddim} and Null-text inversion~\cite{nti}.
For text-based image editing, SDEdit~\cite{sdedit}, Prompt-to-Prompt (P2P)~\cite{p2p}, Plug-and-Play (PnP)~\cite{pnp}, Diffusion Disentanglement (Disentangle)~\cite{disentanglement}, Pix2Pix-Zero~\cite{pix2pixzero} and Cycle Diffusion~\cite{cyclediffusion} are chosen as SOTA approaches. For drag-based image editing, we compare Smooth Diffusion with Stable Diffusion within the framework of DragDiffusion~\cite{dragdiffusion}.






\vspace{2mm} 
\noindent\textbf{Implementation details.} Smooth Diffusion is trained atop pretrained Stable Diffusion-V1.5~\cite{sd}, using LoRA~\cite{lora} finetuning technique. The UNet of Smooth Diffusion is set as trainable with a LoRA rank of 8, while the VAE and text encoder are frozen. We leverage the LAION Aesthetics 6.5+ as the training dataset, which contains 625K image-text pairs with predicted aesthetics scores of 6.5 or higher from LAION-5B~\cite{laion}. Smooth diffusion is typically trained for 30K iterations with a batch size of 96, 3 samples per GPU, a total of 4 A100 GPUs, and a gradient accumulation of 8. The AdamW~\cite{adam} optimizer is adopted with a constant learning rate of $1\times10^{-4}$ and a weight decay of $1\times10^{-4}$. The ratio parameter $\lambda$ in Eq.~\ref{eq:finalloss} is set to 1.
During inference, the total number of diffusion steps is set to 50 and the classifier-free guidance~\cite{cfg} scale is set to 7.5.


\vspace{2mm} 
\noindent\textbf{Evaluation metrics.} To evaluate the general text-to-image generation performance, we report the popular FID~\cite{fid} and CLIP Score~\cite{clip} on the MS-COCO validation set~\cite{coco}. To assess the latent space smoothness, we propose an interpolation standard deviation (ISTD) as an evaluation metric. In specific, we randomly draw 500 text prompts from the MS-COCO validation set. For each prompt, we sample a pair of Gaussian noises and uniformly interpolate them from one to the other 9 times with mix ratios from 0.1 to 0.9. Fed into diffusion models together with a prompt, we could obtain a total of 11 generated images, 2 from the source Gaussian noises and 9 from the interpolated noises. We calculate the standard deviation of L2 distances between every two adjacent images in the pixel space. Finally, we average the standard deviations over 500 prompts as ISTD. Ideally, a zero value of ISTD indicates that consistent and uniform visual fluctuations in the pixel space for identical fixed-size changes in the latent space, resulting in a smooth latent space.
For image inversion, mean square error (MSE), LPIPS~\cite{lpips}, SSIM~\cite{ssim} and PSNR~\cite{psnr} are adopted to evaluate the image reconstruction capability.

\subsection{Latent Space Interpolation}
\textbf{Qualitative comparison.} 
The most straightforward way to demonstrate the smoothness of the latent space is through the observation of interpolation results between latent noises. In~\cref{fig:main}, we present interpolation comparisons between Smooth Diffusion and Stable Diffusion using real images. To generate these comparisons, we utilize the NTI~\cite{nti} to invert a pair of real images into latent noises $\bm{x}_T$, sharing the same $\{\varnothing_t\}_{t=1}^{T}$. We then perform uniform spherical linear interpolations between latent noises (also known as Stable Diffusion Walk~\cite{sdw}), resulting in 9 intermediate noises with mix ratios from 0.1 to 0.9. Subsequently, we concatenate the 11 images produced from these noises to create an image transition sequence in the figures.

Notably, as highlighted by the red boxes, Stable Diffusion exhibits significant visual fluctuations during the transition. In particular, the interpolated images may introduce new attributes that are unrelated to the source images, \eg, the undesired grasslands in the second row of~\cref{fig:main}. In contrast, our approach, Smooth Diffusion, not only avoids introducing obvious irrelevant attributes in the interpolated images but also ensures that the visual effects change smoothly throughout the transition. Additional interpolation results can be seen in supplementary materials.

In addition to Stable Diffusion,~\cref{fig:main} also includes two other baseline methods for comparison: 1) VAE Interpolation (VAE Inter.), which performs interpolations within the VAE space of Stable Diffusion. However, the results closely resemble pixel-space interpolations, with significant degradation of visual details, particularly in the highlighted red box area. 2) ANID~\cite{anid}, which first adds noise to real images and subsequently denoises the interpolated noisy images using Stable Diffusion. In~\cref{fig:main}, ANID with a 50-step scheduler exhibits highly blurred interpolation results. When ANID operates with a default 200-step scheduler, the blurring can be alleviated, but the quality of the interpolated images remains far from satisfactory.

\begin{table}[h]
\centering
\small
\begin{tabular}{cccc}
    \toprule
    Method & ISTD ($\downarrow$)  & FID ($\downarrow$) & CLIP Score ($\uparrow$)\\
    \midrule
    Stable Diffusion & 38.63 & 12.70 & 31.46 \\
    Smooth Diffusion & \textbf{16.54} & \textbf{12.10} & \textbf{31.54} \\
    \bottomrule
\end{tabular}
\caption{\textbf{Quantitative evaluations of image interpolation and text-to-image generation.} We evaluate Smooth Diffusion and Stable Diffusion~\cite{sd} with ISTD, FID~\cite{fid} and CLIP Score~\cite{clip}. The better results are in bold.}
\vspace{-1mm}
\label{table:qc}
\end{table}

\noindent\textbf{Quantitative comparison.} The goal of Smooth Diffusion is to enhance the latent space smoothness without image generation performance degradation compared to Stable Diffusion. In pursuit of this goal, we employ the ISTD introduced in~\cref{sec:setup} to evaluate the latent space smoothness. Additionally, we utilize FID~\cite{fid} and CLIP Score~\cite{clip} to assess generators' overall performance. The results presented in~\cref{table:qc} demonstrate that Smooth Diffusion significantly outperforms Stable Diffusion in terms of ISTD, indicating a substantial improvement in the latent space smoothness. Furthermore, Smooth Diffusion exhibits superior performance in both FID and CLIP Score, suggesting that the enhancement of latent space smoothness and the overall image generation quality are not mutually exclusive but complement each other when the regularization term is applied with a suitable strength ratio.

\begin{figure}[t]
  \centering
  \includegraphics[width=\linewidth]{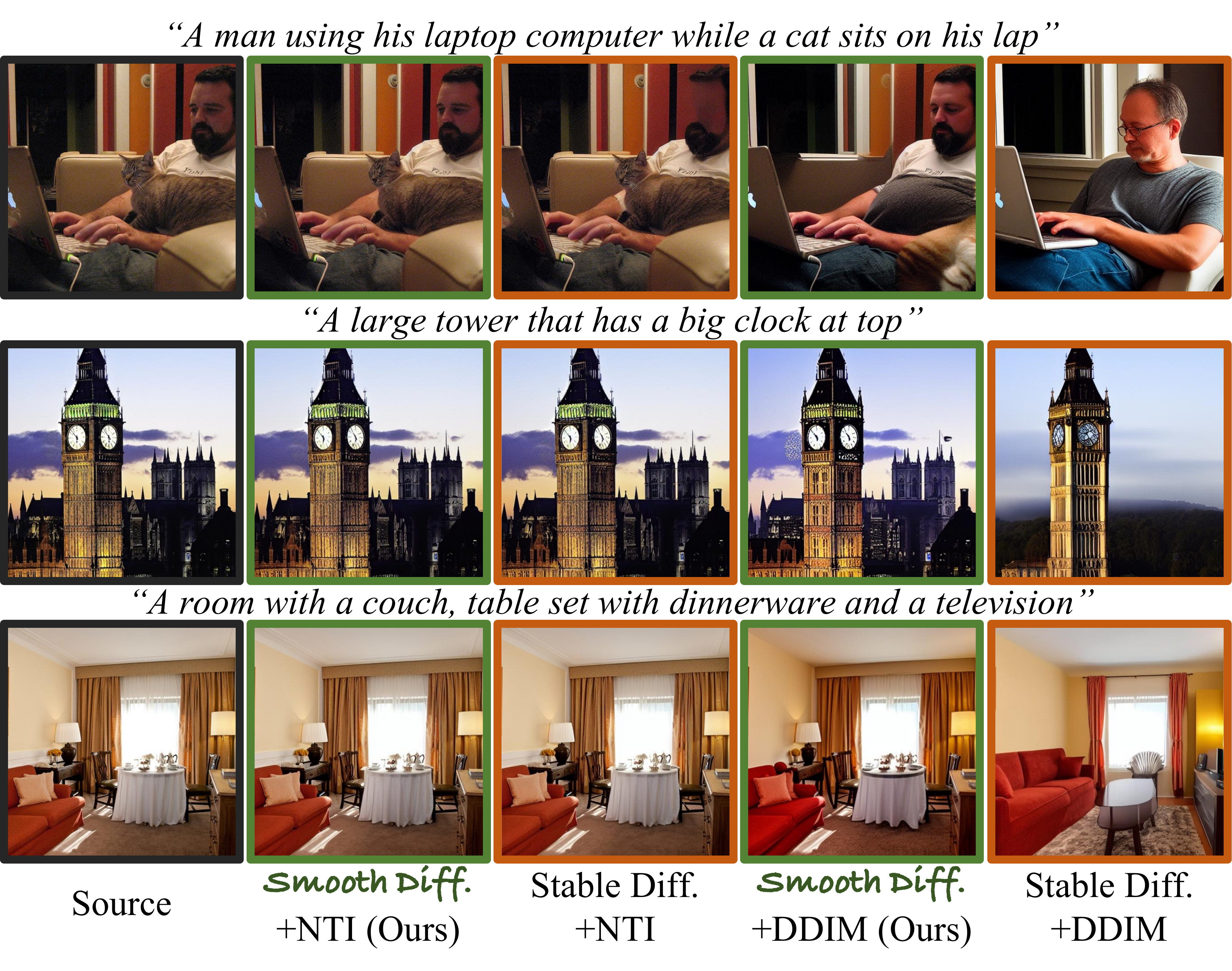}
    \vspace{-8mm}
  \caption{\textbf{Image reconstruction comparison results.} We integrate Smooth Diffusion and Stable Diffusion~\cite{sd} with NTI~\cite{nti} (column 2 \& 3) and DDIM inversion~\cite{ddim} (column 4 \& 5).}
  \label{fig:inversion}
\end{figure}

\subsection{Image Inversion and Reconstruction}
\label{sec:inversion}
Previous research~\cite{sg2} in the realm of GANs discovered that a smoother latent space has a positive impact on the accuracy of image inversion and reconstruction. We empirically validate this finding within the context of diffusion models. 
In specific, two representative inversion techniques, DDIM inversion~\cite{ddim} and Null-text inversion (NTI)~\cite{nti} are adopted and integrated with Smooth Diffusion and Stable Diffusion separately. We both qualitatively and quantitatively compare the image inversion and reconstruction performance of these integrated models using 500 randomly sampled images from the MS-COCO validation set~\cite{coco}. 

\begin{table}[t]
\centering
\resizebox{\linewidth}{!}{
    \begin{tabular}{ccccc}
        \toprule
        Method & MSE ($\downarrow$) & LPIPS ($\downarrow$)& SSIM ($\uparrow$) & PSNR ($\uparrow$)\\
        \midrule         
        Stable Diff. + DDIM & 0.1756 & 0.5385 & 0.2662 & 13.97\\
        Smooth Diff. + DDIM & 0.1086 & 0.4326 & 0.3418 & 16.17\\
        Stable Diff. + NTI  & 0.0156 & 0.1656 & 0.6068 & 25.63\\
        Smooth Diff. + NTI  & \textbf{0.0153} & \textbf{0.1635} & \textbf{0.6102} & \textbf{25.74}\\
        \midrule
        VAE Reconstruction & 0.0148 & 0.1590 & 0.6136 & 25.98\\
        \bottomrule
    \end{tabular}
}
\caption{\textbf{Quantitative evaluations of image reconstruction.} We integrate Stable Diffusion and Smooth Diffusion~\cite{sd} with DDIM inversion~\cite{ddim} (row 2 \& 3) and NTI~\cite{nti} (row 4 \& 5). MSE, LPIPS~\cite{lpips}, SSIM~\cite{ssim} and PSNR~\cite{psnr} are evaluated. VAE Reconstruction results are provided as the optimal values.}
\vspace{-4mm}
\label{table:inversion}
\end{table}

\begin{figure*}[t]
  \centering
  \includegraphics[width=\linewidth]{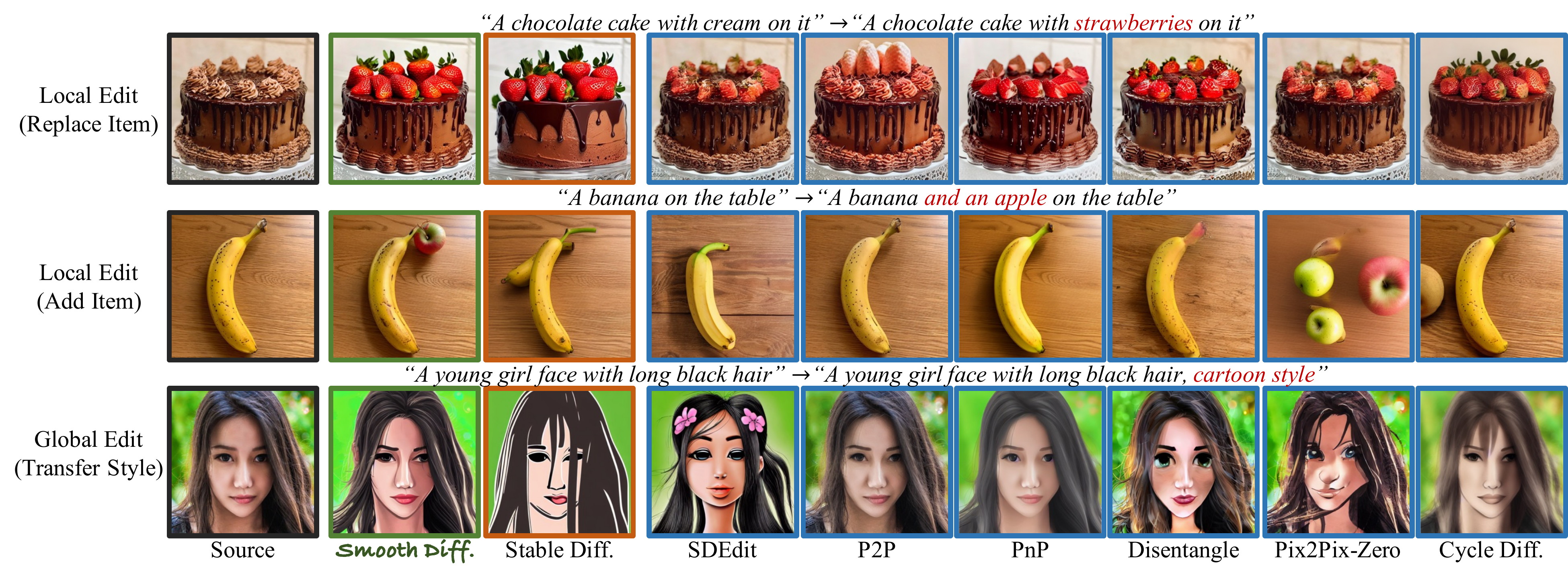}
    \vspace{-7mm}   
  \caption{\textbf{Text-based image editing comparison results.} We compare Smooth Diffusion and Stable Diffusion~\cite{sd} (column 2 \& 3), considering both local and global edits through the straightforward pipeline described in~\cref{sec:editing}. Additionally, we present results from SOTA approaches, including SDEdit~\cite{sdedit}, P2P~\cite{p2p}, PnP~\cite{pnp}, Disentangle~\cite{disentanglement}, Pix2Pix-Zero~\cite{pix2pixzero}, and Cycle Diffusion~\cite{cyclediffusion}, as references.}
  \label{fig:textedit}
  \vspace{-4mm}
\end{figure*}


As illustrated in the two rightmost columns of~\cref{fig:inversion}, when employing a straightforward DDIM inversion, Smooth Diffusion outperforms Stable Diffusion by a considerable margin in terms of reconstruction quality. 
This improvement is evident in various aspects, such as an accurate generation of character identities, a faithful recreation of the city view behind the tower, and a correct reproduction of room layouts.
This phenomenon underscores the fact that the latent space of Smooth Diffusion is more tolerant of the errors introduced by the local linear approximation in DDIM inversion. Consequently, the reconstruction results produced by Smooth Diffusion manage to retain the contents of the source images to a greater extent. On the other hand, when the optimization-based NTI technique is employed, the disparity between Smooth Diffusion and Stable Diffusion is not as pronounced. 
Nonetheless, there are still instances where Stable Diffusion exhibits subpar results, such as the ruined man's face in~\cref{fig:inversion}.

To quantify the image reconstruction performance, MSE, LPIPS~\cite{lpips}, SSIM~\cite{ssim} and PSNR~\cite{psnr} are reported in~\cref{table:inversion}. Notably, the reconstruction error encompasses two components: 1) the error from different inversion methods and U-Net parameters and 2) the error from the shared pretrained VAE~\cite{vae}. Hence, we included the VAE reconstruction errors as optimal values for our method.
The results exhibit a consistent outperformance of Smooth Diffusion over Stable Diffusion across all metrics, whether using DDIM inversion or NTI. Moreover, ``Smooth Diffusion + NTI" performs results close to VAE reconstruction, indicating its superiority attributed to a smoother latent space.



\subsection{Image Editing}\label{sec:editing}
The superiority of Smooth Diffusion in image inversion and reconstruction has motivated us to explore its potential for enhancing image editing tasks. In this section, we delve into two typical image editing scenarios: text-based image editing and drag-based image editing.

\vspace{2mm}

\noindent\textbf{Text-based image editing.} There have been numerous methods~\cite{sdedit,p2p,pnp,disentanglement,pix2pixzero,cyclediffusion} proposed in the literature, each with its own unique designs aimed at achieving the SOTA performance. In contrast, we adopt a simpler pipeline akin to the image inversion and reconstruction process discussed in~\cref{sec:inversion}. The key distinction lies in our approach to modify the text prompt during the later time steps of the reconstruction process. In specific, the original ${\epsilon_\theta}(\bm{x}_t, t, \mathcal{C},\varnothing_t)$ in~\cref{eq:cfg} during NTI reconstruction (diffusion sampling) process is replaced with:
\begin{equation}
\setlength{\abovedisplayskip}{6pt}
\setlength{\belowdisplayskip}{6pt}
{\epsilon_\theta}(\bm{x}_t, t, \mathcal{C}, \varnothing_t)=\left\{\begin{aligned} &{\epsilon_\theta}(\bm{x}_t, t, \mathcal{C}_{\rm src}, \varnothing_t), \ t > T\times r, \\
&{\epsilon_\theta}(\bm{x}_t, t, \mathcal{C}_{\rm trg}, \varnothing_t),\ t \leq T\times r,
\end{aligned}\right.\end{equation}
where $\mathcal{C}_{\rm src}$ represents the source text prompt for inversion, while $\mathcal{C}_{\rm trg}$ corresponds to the target text prompt for editing. The parameter $r$ serves as a threshold, determining when to switch from $\mathcal{C}_{\rm src}$ to $\mathcal{C}_{\rm trg}$. In practice, $r$ is typically chosen within \{0.6, 0.7, 0.8, 0.9\}, with the exact value depending on the specific input images and target visual effects.

\begin{figure}[t]
  \centering
  \includegraphics[width=0.9\linewidth]{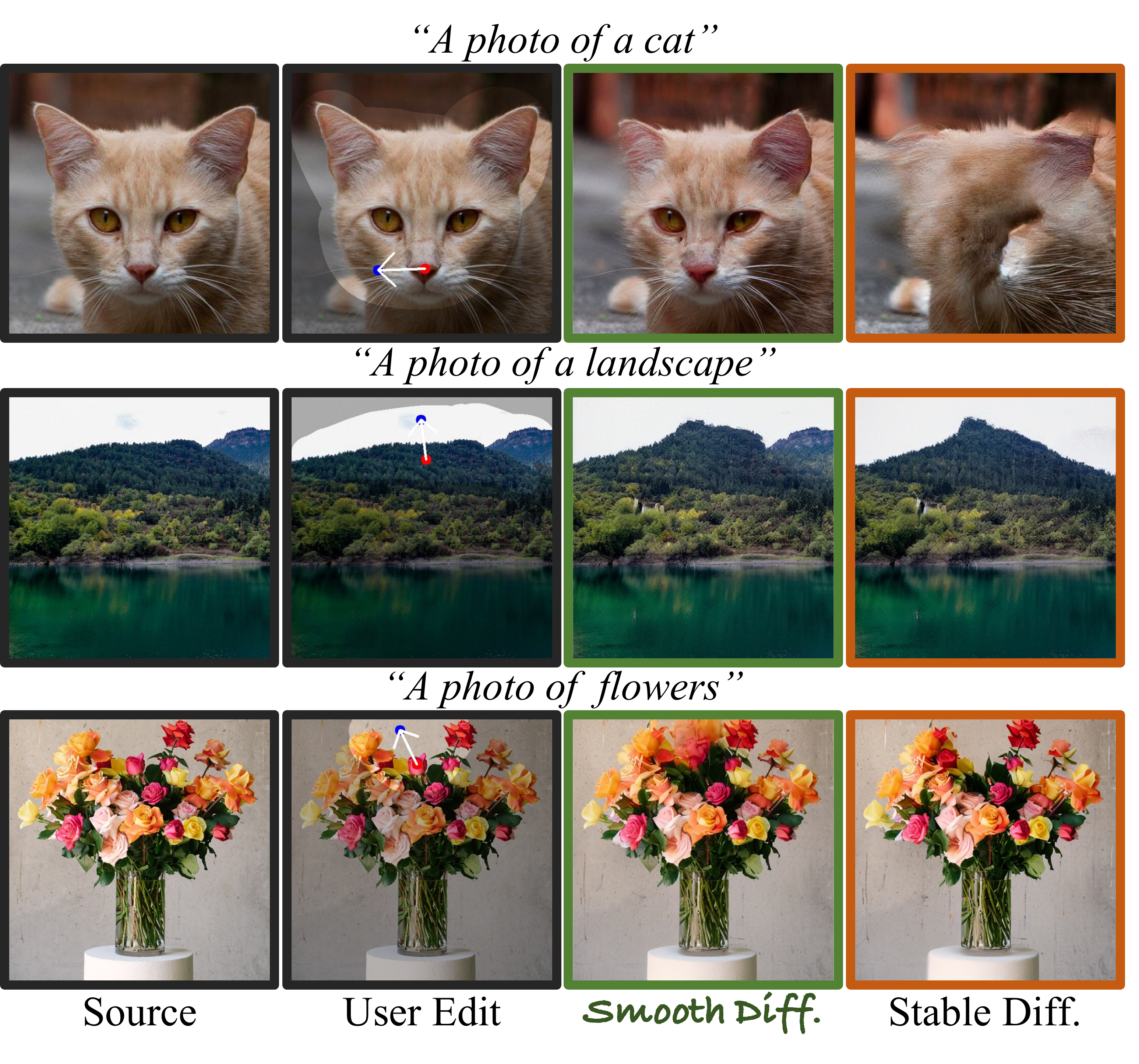}
    \vspace{-3mm}   
  \caption{\textbf{Drag-based image editing comparison results.} We implement Smooth Diffusion and Stable Diffusion~\cite{sd} within the framework of DragDiffusion~\cite{dragdiffusion}, respectively.}
  \label{fig:drag}
  \vspace{-5mm}
\end{figure}

Through this straightforward pipeline, we conducted a comparative analysis of the editing performance between Smooth Diffusion and Stable Diffusion, as presented in the three left-most columns of~\cref{fig:textedit}. We also included editing results obtained from SOTA methods as references. Our evaluation encompasses both local and global editing tasks. The local editing tasks involve replacing items (\eg, changing ``cream" to ``strawberries") and adding items (\eg, ``apple"). On the other hand, the global editing tasks pertain to global style transfer, such as transforming an image into a ``cartoon style". It is evident that while Stable Diffusion excels in achieving precise image reconstruction with NTI, as discussed in~\cref{sec:inversion}, even minor modifications to the text prompt can significantly impact the content of the generated images. For instance, it can affect elements like the style of the cake, the shape of the banana, and the haircut of the girl. In contrast, Smooth Diffusion not only accurately generates edited images in accordance with the target text prompts but also effectively preserves the unedited contents. Furthermore, when compared to SOTA methods, even with this straightforward pipeline, Smooth Diffusion consistently delivers competitive results across all cases.

\vspace{2mm} 
\noindent\textbf{Drag-based image editing.} As an emerging research avenue in the community, drag-based image editing~\cite{draggan,dragdiffusion,dragondiffusion} has garnered considerable attention recently. DragDiffusion~\cite{dragdiffusion} first introduces a framework for drag-based image editing employing Stable Diffusion. In the task 3 of~\cref{fig:fig1} and~\cref{fig:drag}, we showcase that by integrating Smooth Diffusion into the DragDiffusion framework, some previously unsuccessful editing operations with Stable Diffusion can be enabled. As illustrated, Smooth Diffusion achieves operations such as making the tree grow taller without damaging existing branches (\cref{fig:fig1}), rotating the cat head, creating a new mountain top without destroying the original one, and letting new flowers grow in the vase (\cref{fig:drag}). These operations, however, fail with Stable Diffusion, indicating the non-smoothness of its latent space.

\subsection{Ablation Studies}

\textbf{Regularization ratio.} In~\cref{table:ratio}, we examine the impact of different strength ratios $\lambda$ in~\cref{eq:finalloss}. This ratio adjusts the intensity of the step-wise variation regularization. Specifically, when a weaker regularization is applied (\eg, $\lambda=0.1$), we observe a slight improvement in the CLIP Score. However, there is a significant increase in ISTD, indicating a notable degradation in latent space smoothness. In contrast, employing a stronger regularization (\eg, $\lambda=10$) leads to a smoother latent space, as demonstrated by the decrease in ISTD. However, in this case, we observe an unexpected increase in FID, indicating a notable decline in the quality of generated images. Therefore, selecting an appropriate trade-off value for $\lambda$ becomes crucial based on the specific experimental settings. In our default setting, we find that $\lambda=1$ serves as a suitable value.

\begin{table}[h]
\centering
\small
\begin{tabular}{cccc}
    \toprule
    Ratio& ISTD ($\downarrow$) & FID ($\downarrow$) & CLIP Score ($\uparrow$)\\
    \midrule
    0.1 & 24.23 & \underline{12.15} & \textbf{31.56}\\
     1 (default) & \underline{16.54} & \textbf{12.11} & \underline{31.49}\\
    10 & \textbf{11.51} & 17.44 & 31.41 \\
    \bottomrule
\end{tabular}
\vspace{-1mm} 
\caption{\textbf{Ablation results of different regularization ratios.} The best results are in bold, and the second-best results are underlined.}
\vspace{-2mm} 
\label{table:ratio}
\end{table}

\noindent\textbf{LoRA rank.} In~\cref{table:rank}, we examine the impact of different ranks of the LoRA component utilized in our Smooth diffusion. We discover that LoRA ranks within the range of [4,16] are all suitable values for our default setting. We select a default rank of 8 because of its lowest ISTD among the first three rows in~\cref{table:rank}. Furthermore, we train a fully finetuned model, referred to as "full," which showcases a further decrease in ISTD. However, this comes at the expense of significantly degrading the quality of the generated images, as indicated by an increased FID and decreased CLIP Score. This decline in performance underscores the vulnerability of fully fine-tuned models to collapse within our default setting, emphasizing the need for additional meticulous design considerations.

\begin{table}[h]
\centering
\small
\begin{tabular}{cccc}
    \toprule
    Rank& ISTD ($\downarrow$) & FID ($\downarrow$) & CLIP Score ($\uparrow$)\\
    \midrule
    4 & 16.76&12.36&31.49\\
    8 (default)&\underline{16.54}&\underline{12.11}&\underline{31.54}\\
    16&16.65&\textbf{11.49}&\textbf{31.61}\\
    full&\textbf{11.52}&27.27&28.86 \\
    \bottomrule
\end{tabular}
\vspace{-1mm} 
\caption{\textbf{Ablation results of different LoRA ranks.} The best results are in bold, and the second-best results are underlined.}
\vspace{-4mm} 
\label{table:rank}
\end{table}

\section{Conclusion}

\label{sec:con}
In this article, we explored Smooth Diffusion, an innovative diffusion model that enhances latent space smoothness for generation. Smooth Diffusion adopts the novel Step-wise Variation Regularization, which successfully maintains variation between arbitrary input latent and generated images at a more bounded range. Smooth Diffusion was trained on top of the prevailing text-to-image model, from which we carried out extensive research, including but not limited to interpolation, inversion, and editing, all of which had shown competitive performance. Through qualitative and quantitative measurements, we demonstrated that Smooth Diffusion managed to make a smoother latent space without compromising the output quality. We believe that Smooth Diffusion will become a valuable solution for other challenging tasks, such as video generation, in the future.


{
    \small
    \bibliographystyle{ieeenat_fullname}
    \bibliography{main}
}

\clearpage
\appendix
\section*{Supplementary Materials}
\section{Implementation Details}

This section elaborates on details briefly introduced in the main paper. These include the notation, the basic training objective, the interpolation standard deviation (ISTD) metric, and our utilization of Null-text inversion (NTI)~\cite{nti} for real-image interpolation.

\subsection{Notation} Stable Diffusion~\cite{sd} employs an efficient ``latent'' diffusion pipeline. Here the ``latent'' refers to using an individually trained (VAE)~\cite{vae} to compress an input image $\bm{x}_0$ into its VAE-space representation $\bm{z}_0$:
\begin{equation}
\setlength{\abovedisplayskip}{5pt}
\setlength{\belowdisplayskip}{5pt}{
\begin{aligned}
    \bm{z}_0 = \mathcal{E}(\bm{x}_0), \quad \bm{x}_0=\mathcal{D}(\bm{z}_0),
\end{aligned}}\end{equation}
where $\mathcal{E}$ and $\mathcal{D}$ represent the encoder and decoder of the VAE, respectively. For simplicity, we exclude this conversion process and only use ``$\bm x$''-based notations in the main paper. Although we chose Stable Diffusion as our baseline due to its popularity and high performance, our training pipeline is not specifically tailored for latent diffusion models and is compatible with other diffusion models.


\subsection{Basic Training Objective} Smooth Diffusion's training objective comprises two key components: 1) a basic training objective primarily centered on noise prediction but flexible in formulation for different diffusion models, and 2) our proposed Step-wise Variation Regularization term. In our experiments, the basic training objective is:
\begin{equation}
\setlength{\abovedisplayskip}{5pt}
\setlength{\belowdisplayskip}{5pt}{
\begin{aligned}
    \mathcal{L}_{\rm{base}} = \mathbb{E}_{\bm{x}_0,\bm{\epsilon},t}\Vert\bm{\epsilon}-{\epsilon_\theta}(\bm{x}_t, t)\Vert_2^2,
    \label{eq:base}
\end{aligned}}\end{equation}
which is a commonly adopted training objective across many diffusion models, \eg, Stable Diffusion~\cite{sd}.

\subsection{ISTD} The goal of ISTD is to quantify the deviation of pixel-space changes given the same fixed-step changes in latent space. A lower deviation implies the input latents and output images are more likely to change smoothly. In our experiments, we first randomly draw 500 text prompts from the MS-COCO validation set~\cite{coco}. For each prompt, we then sample two random Gaussian noises, $\bm{\epsilon}^a$ and $\bm{\epsilon}^b$.
Next, we execute uniform spherical linear interpolations (slerp) between $\bm{\epsilon}^a$ and $\bm{\epsilon}^b$ for 11 times, varying the mixing ratio $\eta$ from 0 to 1:
\begin{equation}
\setlength{\abovedisplayskip}{5pt}
\setlength{\belowdisplayskip}{5pt}{
\begin{aligned}
    \bm{\epsilon}^{\eta} = {\rm slerp}(\bm{\epsilon}^a, \bm{\epsilon}^b, \eta),\quad \eta = 0, 0.1, 0.2, \cdots, 1.
    \label{eq:supp_inter}
\end{aligned}}\end{equation}

We employ the testing diffusion model to generate 11 interpolated images $\{\widehat{\bm{x}_0^{\eta}}\}_{\eta=0}^{1}$ from $\{\bm{\epsilon}^{\eta}\}_{\eta=0}^{1}$. Notice that Eq. \ref{eq:supp_inter} guarantees that the latent space changes between every two adjacent latents (\ie, $\bm{\epsilon}^{\eta}$ and $\bm{\epsilon}^{\eta + 0.1}$) are the same. Hence, we calculate the L2 distances between every two adjacent images (\ie, $\widehat{\bm{x}_0^{\eta}}$ and $\widehat{\bm{x}_0^{\eta + 0.1}}$ ) and compute the standard deviation of these distances. Finally, ISTD is the average of standard deviations over 500 different text prompts. For a fair comparison, the text prompts and the noises for each prompt are the same for different testing models.

\subsection{NTI for real-image interpolation} NTI is initially designed to transform a real image $\bm{x}_0$ into a latent $\widetilde{\bm{x}_T}$, along with a series of learnable null-text embeddings $\{\varnothing_t\}_{t=1}^{T}$ for each step $t$. The optimization for each $\varnothing_t$ is formulated as:
\begin{equation}
\setlength{\abovedisplayskip}{6pt}
\setlength{\belowdisplayskip}{6pt}{
\begin{aligned}
    \min_{\varnothing_t}\Vert \widetilde{\bm{x}_{t-1}}- {\rm DDIM}(\widetilde{\bm{x}_t}, t, \xi, \varnothing_t)\Vert_2^2.
\end{aligned}}\end{equation}
where $\{\widetilde{\bm{x}_{t}}\}_{t=1}^{T}$ represents intermidiate noisy images estimated by DDIM inversion~\cite{ddim}. For simplicity, ${\rm DDIM}(\widetilde{\bm{x}_t}, t, \xi, \varnothing_t)$ denotes the DDIM sampling process at step $t$, utilizing the text embedding $\xi$, the null-text embedding $\varnothing_t$ and the classifier-free guidance scale $w=7.5$.

For real-image interpolation, we optimize a shared series of $\{\varnothing_t\}_{t=1}^{T}$ for two real images, $\bm{x}_0^a$ and $\bm{x}_0^b$:
\begin{equation}
\setlength{\abovedisplayskip}{6pt}
\setlength{\belowdisplayskip}{6pt}{
\begin{aligned}
    \min_{\varnothing_t}\Vert \widetilde{\bm{x}_{t-1}^a}- {\rm DDIM}(\widetilde{\bm{x}_t^a}, t, \xi, \varnothing_t)\Vert_2^2 +\\ \Vert \widetilde{\bm{x}_{t-1}^b}- {\rm DDIM}(\widetilde{\bm{x}_t^b}, t, \xi, \varnothing_t)\Vert_2^2.
\end{aligned}}\end{equation}

In our experiments, we only interpolate the latents $\widetilde{\bm{x}_{T}^a}$ and $\widetilde{\bm{x}_{T}^b}$ following Eq. \ref{eq:supp_inter} and use the same null-text embeddings $\{\varnothing_t\}_{t=1}^{T}$ for all interpolated images.




\section{Additional Results}

\begin{figure*}[t]
  \centering
  \includegraphics[width=\linewidth]{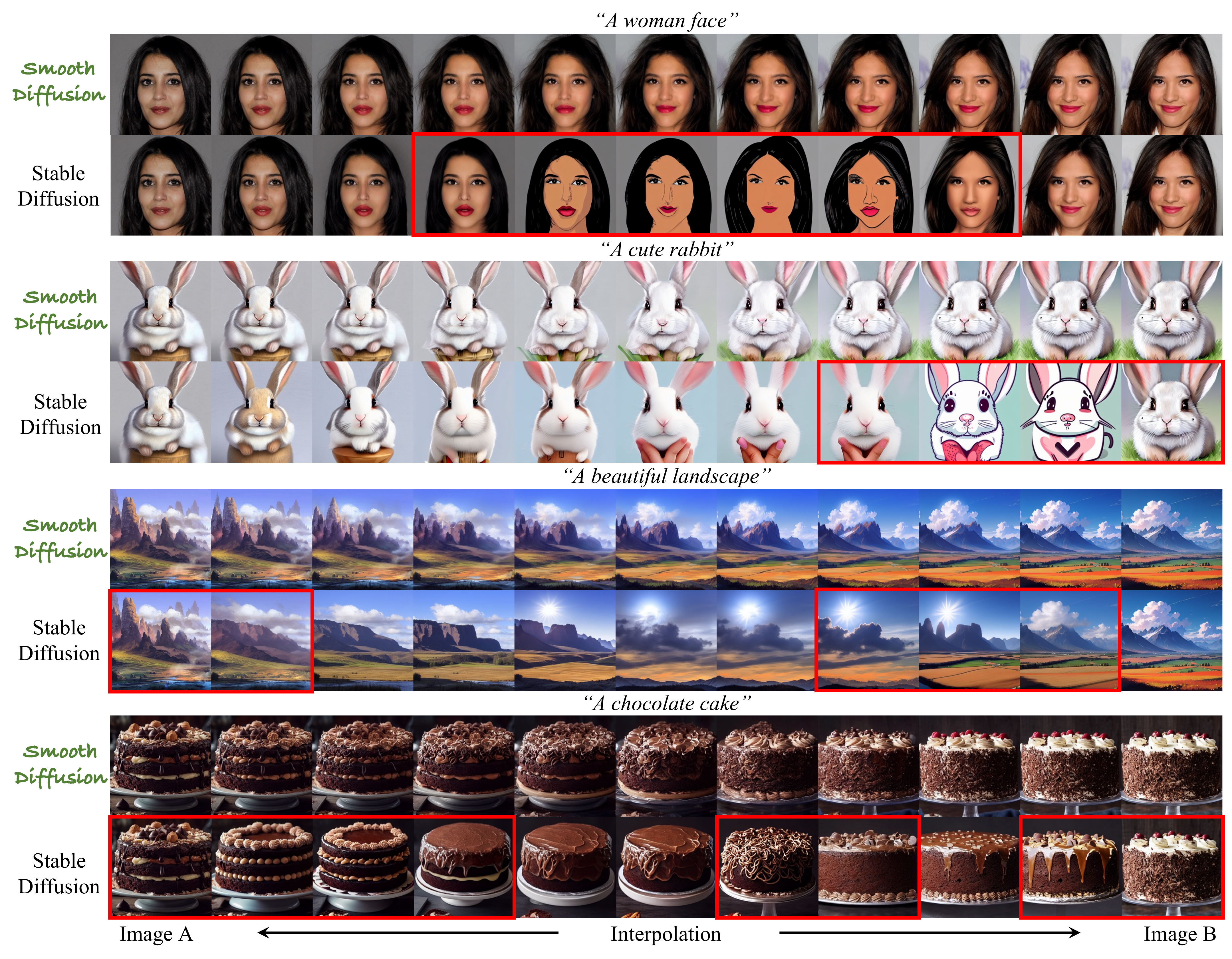}
  \caption{\textbf{Additional image interpolation results with Smooth Diffusion.} For Smooth Diffusion and Stable Diffusion~\cite{sd}, real images (Image A and B) are inverted into latents using Null-text inversion~\cite{nti}. We perform spherical linear interpolations between latents and concatenate the resulting images as a transition sequence.}
  \label{fig:supp_inter}
\end{figure*}

\begin{figure*}[t]
  \centering
  \includegraphics[width=\linewidth]{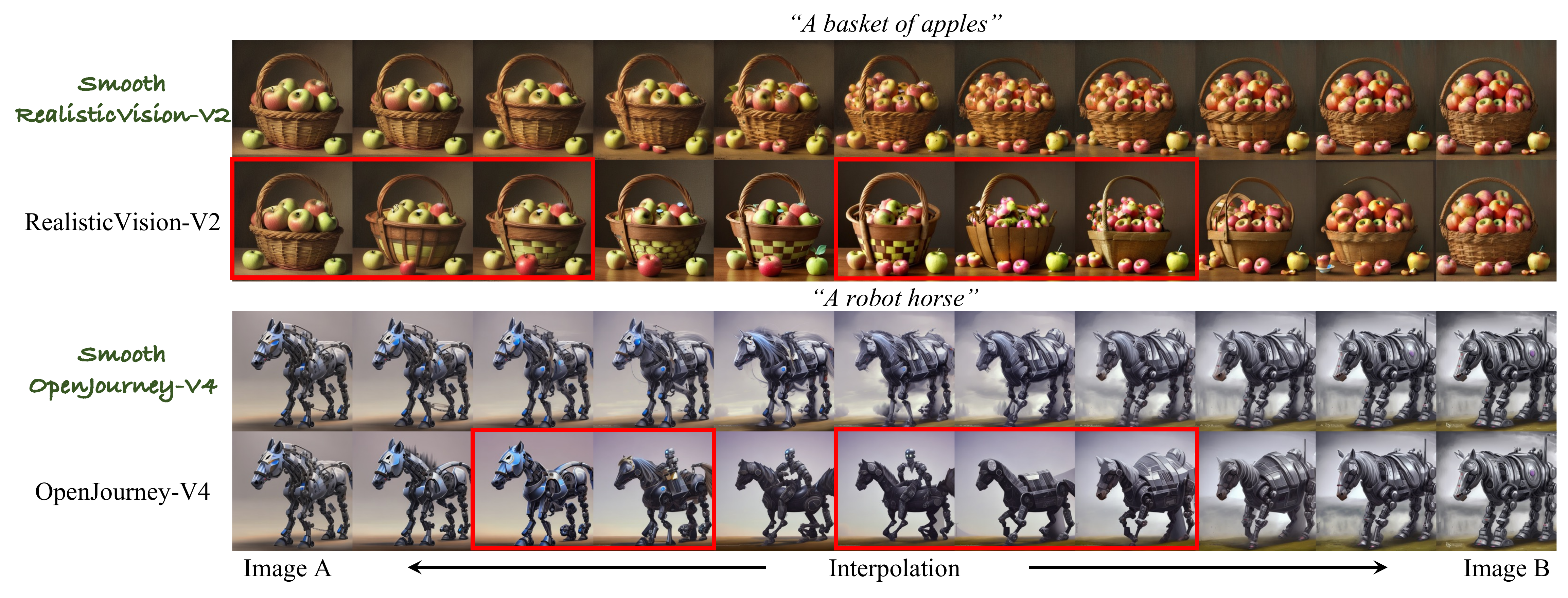}
  \caption{\textbf{Image interpolation results with community models.} We apply the LoRA component of Smooth Diffusion to RealisticVision-V2~\cite{rvv2} and OpenJournery-V4~\cite{ojv4} and perform spherical linear interpolations in their latent spaces.}
  \label{fig:supp_reuse}
\end{figure*}

\begin{figure*}[t]
  \centering
  \includegraphics[width=\linewidth]{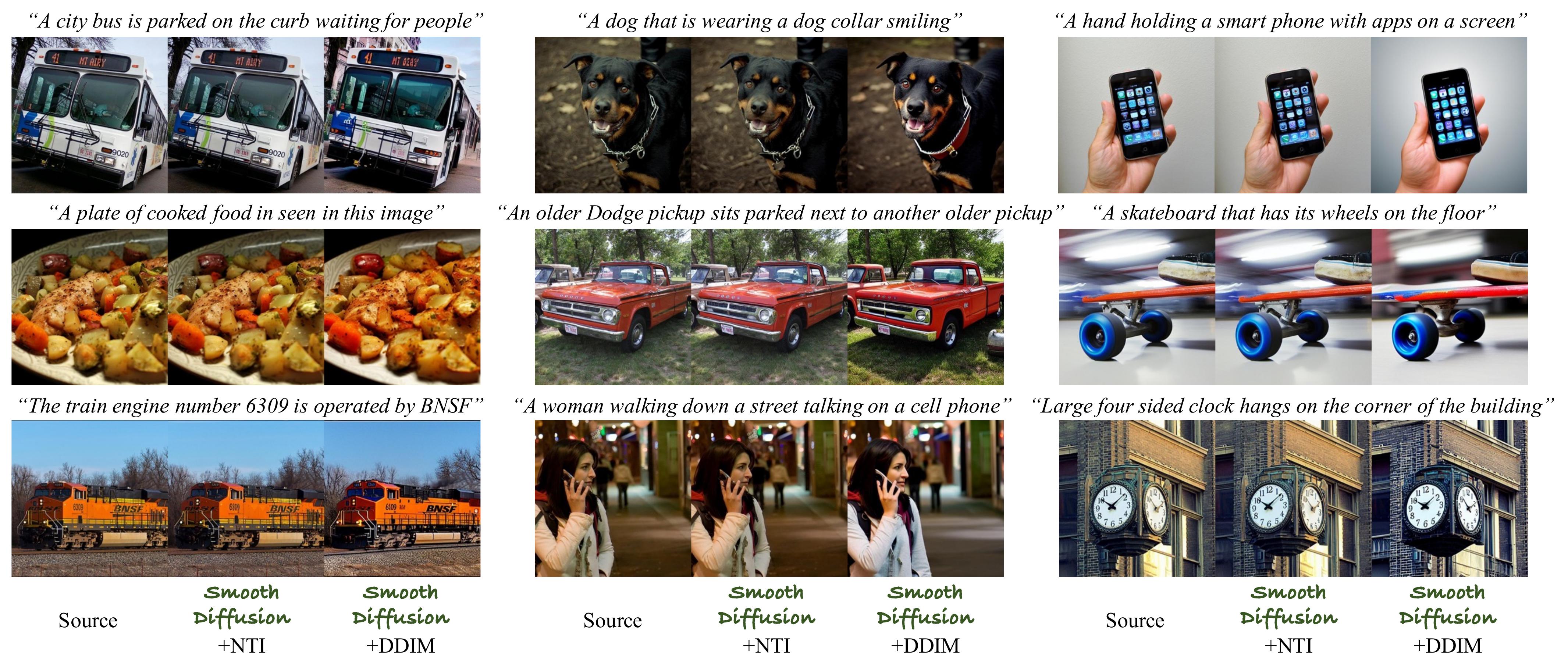}
  \caption{\textbf{Additional image inversion and reconstruction results with Smooth Diffusion.} We integrate Smooth Diffusion with two typical diffusion inversion techniques, Null-text inversion~\cite{nti} and DDIM inversion~\cite{ddim}.}
  \label{fig:supp_inv}
\end{figure*}

\begin{figure*}[t]
  \centering
  \includegraphics[width=\linewidth]{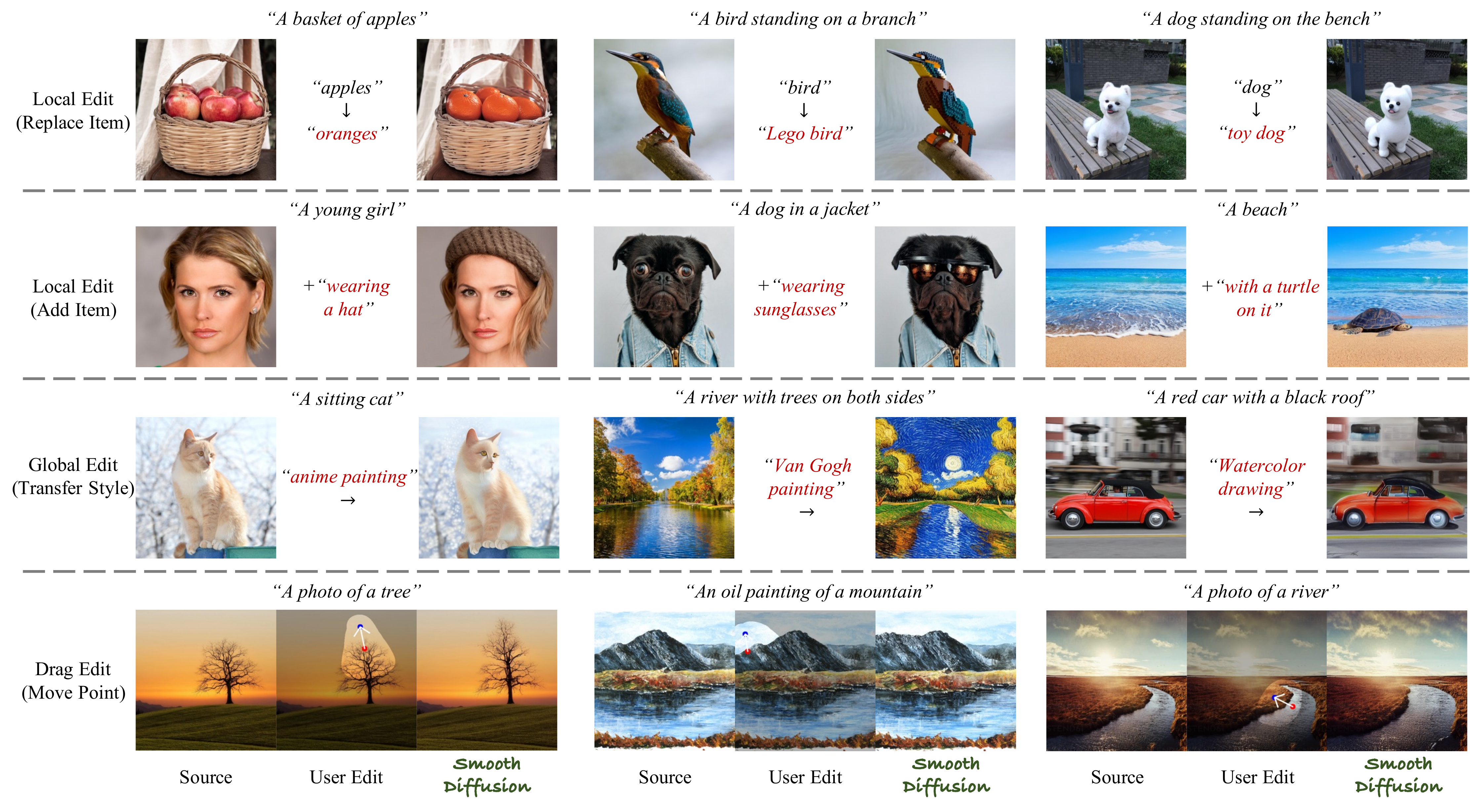}
  \caption{\textbf{Additional image editing results with Smooth Diffusion.} Both text-based image editing and drag-based image editing are evaluated. For text-based image editing, we consider both local and global edits to test Smooth Diffusion. For drag-based image editing, Smooth Diffusion is integrated into the framework of DragDiffusion~\cite{dragdiffusion}.}
  \label{fig:supp_edit}
\end{figure*}

This section provides additional visual results of Smooth Diffusion. We display image interpolation results in~\cref{fig:supp_inter} and~\cref{fig:supp_reuse}, image inversion and reconstruction results in~\cref{fig:supp_inv}, and image editing results in~\cref{fig:supp_edit}.

\textbf{Reusability.} 
The LoRA component of Smooth Diffusion remains adaptable to other models sharing the same architecture as Stable Diffusion. However, the effectiveness of this reusability is not guaranteed. We evaluate the integration of this LoRA component into two popular community models, RealisticVision-V2~\cite{rvv2} and OpenJourney-V4~\cite{ojv4}. As depicted in~\cref{fig:supp_reuse}, this integration also enhances the latent space smoothness of these models. This reusability makes our method eliminate the need for repeated training and become a plug-and-play module across various models.

\end{document}